\documentclass[conference]{IEEEtran}
\usepackage{stmaryrd}
\usepackage{amsfonts}
\usepackage{graphicx,times,psfig,amsmath}

\usepackage{graphicx}
\usepackage{amsmath}
\usepackage{amssymb}
\usepackage{algorithmic}
\usepackage{algorithm}
\usepackage{subfigure}
\usepackage{array}
\usepackage{multirow,tabularx}
\usepackage{multicol}

\newcommand{\bfx}{{\textbf{x}}}
\newcommand{\bfv}{{\textbf{v}}}
\newcommand{\bfu}{{\textbf{u}}}

\newcommand{\bfmu}{{\boldsymbol{\mu}}}

\newcommand{\bfalpha}{{\boldsymbol{\alpha}}}
\newcommand{\bfbeta}{{\boldsymbol{\beta}}}

\IEEEoverridecommandlockouts

\begin{document}

\title{\ \\ \LARGE\bf
Large Margin Image Set Representation and Classification
\thanks{Jim Jing-Yan Wang  is  with the
University at Buffalo, The State University of New York,
Buffalo, NY 14203, USA.}
\thanks{Majed Alzahrani  and Xin Gao are with the
Computer, Electrical and Mathematical Sciences and Engineering Division,
King Abdullah University of Science and Technology (KAUST),
Thuwal 23955-6900, Saudi Arabia.}}

\author{Jim Jing-Yan Wang, Majed Alzahrani and Xin Gao}

\maketitle

\begin{abstract}
In this paper, we propose a novel image set representation and classification method by maximizing the
margin of image sets.
The margin of an image set is defined as
the difference of the distance to its nearest
image set from different classes
and the distance to its nearest image set of the same class.
By modeling the image sets by using both their
image samples and their affine hull models,
and maximizing the margins of the images sets,
the image set representation parameter learning problem is formulated as an
minimization problem, which is further optimized by an expectation---maximization (EM) strategy
with accelerated proximal
gradient (APG) optimization
in an iterative algorithm.
To classify a given test image set, we assign it to the class which could provide the
largest margin.
Experiments on two applications of video-sequence-based face recognition
demonstrate that the proposed method significantly outperforms state-of-the-art
image set classification methods in terms of both effectiveness and efficiency.
\end{abstract}

\section{Introduction}

\PARstart{T}{raditional}
visual information processing and understanding is based on single images.
The classification of a visual target, such as a human face, is also conducted based on a single image,
where each training or test sample is an individual image \cite{Zhao2003399,JonathonPhillips20001090,wang2012scimate,wang2012adaptive,zhou2010region}.
With the rapid development of video technologies,
sequences of images is more commonly available than single images.
Consequently, the visual target classification task could be
improved from image-based to image-set-based.
The image set classification problem has been proposed and studied recently
\cite{MMD2008,Chu20111567,Fan2006177,Kim2007,Chin2006461,Wang20091161,MMD2008,Arandjelovic2005581,sun2012unsupervised}.
In image set classification,
each sample is a set of images instead of one single image, and
each class is represented by one or
more training samples.
The classification problem is to assign a given test sample
to one of the known classes.
For example, in image-set-based face recognition problem,
each sample is a set of facial images with different poses, illuminations,
and expressions.
Compared to the traditional single image classification,
image set classification has the potential of achieving higher accuracy,
because image sets usually contain more information than single images.
Even if images in the image set are of low quality,
we could still exploit the
temporal
relationship and the complementarity between the images to improve the classification
accuracy.

Although image set classification has proposed a novel and promising scheme for the visual
classification problem, it has also brought challenges to the machine learning and
computer vision communities.
Traditional single-image-based representation and classification
methods are not suitable for this problem,
such as
principal component analysis (PCA) \cite{Wold198737},
support vector machine (SVM) \cite{SVM2013}, and K-nearest neighbors (KNN) \cite{KNN2013}.
To handle the problem of image set classification, a number of methods have been proposed.
In \cite{Kim2007},
Kim et al. developed
the
discriminant canonical correlations (DCC) for image set classification, by
proposing a linear discriminant function to simultaneously maximize the canonical correlations of within-class sets and minimize the canonical correlations of between-class sets.
The classification is done by transforming the image sets using the discriminant function
and comparing them by the canonical correlations.
In \cite{MMD2008},
Wang et al.
formulated the image set classification problem as the computation of manifold-manifold distance (MMD), i.e., calculating the distance between nonlinear manifolds, each of represented one image set.
In \cite{MDA2009}, Wang et al.
presented the manifold discriminant analysis (MDA) for the image set classification problem, by modeling each image set
as a manifold and
formulating the problem as a classification-oriented multi-manifold learning problem.
Cevikalp and Triggs later
introduced the affine hull-based image set distance (AHISD) for image set based face recognition \cite{AHISD2010},
by representing images as points in a linear or affine feature space and characterizing each image set by
a convex geometric region spanned by its feature points.
Set dissimilarity was measured by geometric distances between convex models.
Hu et al.
proposed to
represent each image set as both the
image samples of the set and their affine hull model \cite{Hu2011,Hu2012},
and
introduced a novel between-set distance called sparse approximated
nearest point (SANP) distance for the image set classification.
In \cite{Wang2012},
Wang et al.
modeled the image set with its
natural second-order statistic, i.e. covariance matrix (COV),
and
proposed a novel discriminative learning approach to
image set classification based on the covariance matrix.

Among all these methods, affine subspace-based methods, including AHISD \cite{AHISD2010}
and SANP \cite{Hu2011,Hu2012}, have shown their advantage over the other methods.
However, all these methods are unsupervised ones, ignoring the class labels of the
images sets in the training set.
Moreover, most image-set-based classification methods are
under the framework of pairwise image set comparison \cite{MMD2008,MDA2009,AHISD2010,Hu2011,Hu2012}. A  similar approach has been successfully adopted on pairwise comparison on individual samples by Mu et al. \cite{mu2013local}, which exploits abundant discriminative information for each local manifold.
In set classification case,  a test image set is compared to all
the training image set one by one, and
then the nearest neighbor rule is utilized to decide which class the test image set belongs to.
The disadvantage of this strategy lays in the following two folds:
\begin{itemize}
\item When the training image set number is large, this strategy is quite time-consuming.
\item When a pair of image sets are compared, all other image sets are ignored,
and thus the global structure of the image set dataset is ignored.
\end{itemize}
To overcome these issues, in this paper,
we propose a novel image set representation and classification method.
Similarly to SANP \cite{Hu2011,Hu2012}, we also use the image samples of an image set and
their affine hull model to represent the image set.
To utilize the class labels of each image set, inspired by large margin framework for feature selection \cite{sun2010local}, we propose to
maximize the margin of each image set.
Based on this representation and its corresponding pairwise distance measure,
we define two types of nearest neighboring image sets for each image set ---
the nearest neighbor from the same class and the nearest neighbor from different classes.
The margin of a image set is defined as the difference between its
distances to nearest miss and nearest hit,
and the representation parameter is learned by maximizing the
margins of the image sets.
To classify a test image set, we assign it to the
class which could achieve the largest margin for it.
The contributions of the proposed Large Margin Image Set (LaMa-IS) representation
and classification method are of three folds:

\begin{enumerate}
\item Using the class labels, we define the margin of the image sets,
such that the discriminative ability can be improved in a supervised manner.
\item The global structure of the image sets can also be
explored by searching the nearest hit and nearest miss from the
entire database for each images set.
\item To classify a test image set, we only need to compare it to every class
instead of every training image set, which could reduce the time complexity of the
online classification procedure significantly, especially when the number of
training image sets is much larger than the number of classes.
\end{enumerate}

The rest of the paper is organized as follows:
in Section \ref{sec:Method}, we propose the novel LaMa-IS algorithm;
in Section \ref{sec:experiment}, the experiment results on several
image-set-based face recognition problems are given;
and finally in Section \ref{sec:conclusion}, we draw conclusions.

\section{Methods}
\label{sec:Method}

In this section we will introduce the proposed LaMa-IS
method for image set representation and classification.

\subsection{Objective Function}

Suppose we have a database of image sets denoted as $\{(X_i,y_i)_{i=1}^N\}$,
where $X_i$ is the data matrix of
the $i$-th image set, and $y_i\in \{1,\cdots,C\}$ is its corresponding class label.
In the data matrix $X_i = [\bfx_{i,1},\cdots,\bfx_{i,N_i}]\in \mathbb{R}^{D\times N_i}$, the $n$-th column,
$\bfx_{i,n}\in \mathbb{R}^D$ is the $D$-dimensional visual feature vector of
the $n$-th image of the $i$-th image set, and
$N_i$ is the number of images in $i$-th image set.
Note that the feature vector for an image can be the original pixel values
or some other visual features extracted from the image, such as
local binary pattern (LBP) \cite{LBP2013}.
To represent an image set, two linear model has been employed to
approximate the structure of the image set following \cite{Hu2011,Hu2012}:
\begin{itemize}
\item Using the images in the image set, we can model the $i$-th image set
as an linear combination of the images in the $i$-th set as
\begin{equation}
\begin{aligned}
\bfx = \sum_{n=1}^{N_i} \bfx_{i,n} \alpha_{i,n} =  X_i \bfalpha_i,
\end{aligned}
\end{equation}
where $\bfalpha_i=[\alpha_{i,1},\cdots,\alpha_{i,N_i}]^\top\in \mathbb{R}^{N_i}$ is the linear combination coefficient vector.

\item We can also use the
affine hull model to represent the image set,
using the
image mean
and the
orthonormal bases of the
$i$-th image, which is represented as
\begin{equation}
\begin{aligned}
\bfx =  \bfmu_i + U_i \bfv_i,
\end{aligned}
\end{equation}
where $\bfmu_i=\frac{1}{N_i}\sum_{n=1}^{N_i} \bfx_{i,n}$ is the image mean,
the columns of $U_i$ are the orthonormal bases obtained
from the SVD of the centered $X_i$, and $\bfv_i$ is the coefficient vector of $U_i$.
\end{itemize}
In this paper, we try to represent the image sets using both models mentioned above simultaneously,
by solving the parameters $\bfalpha_i$ and $\bfv_i$.
The representation error is given by the squared $l_2$ norm distance
between these two models as
\begin{equation}
\begin{aligned}
\mathcal{R}_{\bfv_i,\bfalpha_i}=||(\bfmu_i + U_i \bfv_i) - X_i \bfalpha_i||^2_2,
\end{aligned}
\end{equation}
where $\bfalpha_i$ and $\bfv_i$ are the parameters for the two models, respectively.
To compare a pair of image sets, we only use the second model and compute the
squared
$l_2$ norm distance between them  suggested by \cite{Hu2011} as
\begin{equation}
\begin{aligned}
\mathcal{D}_{\bfv_i,\bfv_j}=||(\bfmu_i + U_i \bfv_i) - (\bfmu_j + U_j \bfv_j)||^2_2.
\end{aligned}
\end{equation}

Given the defined distance
function $\mathcal{D}_{\bfv_i,\bfv_j}$,
we can find two nearest neighboring sets for each set $X_i$,
one from the same class as $X_i$ (called nearest hit or $\mathcal{H}_i$) and the
other from different classes (called nearest miss or $\mathcal{M}_i$),
defined as
\begin{equation}
\begin{aligned}
\mathcal{H}_i = \underset{j:y_j = y_i,j\neq i}{argmin}  \mathcal{D}_{\bfv_i,\bfv_j},\\
\mathcal{M}_i = \underset{j:y_j \neq y_i,j\neq i}{argmin}  \mathcal{D}_{\bfv_i,\bfv_j}.
\end{aligned}
\end{equation}
The margin of the $i$-th set is then defined as
the difference of the distances between $\mathcal{M}_i $
and $\mathcal{H}_i$, as
\begin{equation}
\begin{aligned}
\rho_i=
\mathcal{D}_{\bfv_i,\bfv_{\mathcal{M}_i}}
-
\mathcal{D}_{\bfv_i,\bfv_{\mathcal{H}_i}},
\end{aligned}
\end{equation}
where $\mathcal{H}_i^{y_i}$ is the nearest set from the same class of the $i$-th set, and
$\mathcal{M}_i^{y_i}$ is the nearest one form a different class.
The main difficulty here is that $\bfv_i$ is a variable to be solved when we compute the margin. Thus, it is impossible to directly
find the nearest neighbors.
To overcome this problem,
following the principles of
the expectation-maximization (EM) algorithm, 
we develop a probabilistic
model where the nearest neighbors of a given set are
treated as hidden variables.
The probabilities of the $j$-th set being the
nearest miss or hit of the $i$-th set is denoted as
$P(j=\mathcal{M}_i|\{\bfv_i\})$ and $P(j=\mathcal{H}_i|\{\bfv_i\})$, respectively,
and they are  estimated via the standard kernel density estimation:
\begin{equation}
\label{equ:probality}
\begin{aligned}
&P(j=\mathcal{H}_i|\{\bfv_i\})=
\left\{\begin{matrix}
\frac{K_{\bfv_i,\bfv_j}}
{\sum_{k:y_k=y_i} K_{\bfv_i,\bfv_k}},
& if~ y_j = y_i,~and~j\neq i \\
0, &else
\end{matrix}\right.\\
&P(j=\mathcal{M}_i|\{\bfv_i\})=
\left\{\begin{matrix}
\frac{K_{\bfv_i,\bfv_j}}
{\sum_{k:y_k\neq y_i} K_{\bfv_i,\bfv_k}},
& if~ y_j \neq y_i,~and~j\neq i \\
0 &else
\end{matrix}\right.
\end{aligned}
\end{equation}
where $K_{\bfv_i,\bfv_j}=exp(-\frac{\mathcal{D}_{\bfv_i,\bfv_j}}{2\sigma^2})$
is the Gaussian kernel function,
and $\sigma$ is the band-width parameter.
Subsequently the probabilistic
margin is defined as
\begin{equation}
\begin{aligned}
\rho_i=&
\sum_{j=1}^N P(j=\mathcal{M}_i|\{\bfv_i\}) \times \mathcal{D}_{\bfv_i,\bfv_j}\\
&-
\sum_{j=1}^N P(j=\mathcal{H}_i|\{\bfv_i\})\times \mathcal{D}_{\bfv_i,\bfv_j}
\end{aligned}
\end{equation}
Apparently, the margin of each set should be maximized. Considering the representation error to be minimized and the
margin to be maximized simultaneously, we have the following objective function regarding the parameters  $\bfv_i$ and $\bfalpha_i$
as follows
\begin{equation}
\begin{aligned}
\label{equ:objective}
\underset{\{\bfalpha_i,\bfv_i\}}{min}
\sum_{i=1}^N
&
\left \{
\mathcal{R}_{\bfv_i,\bfalpha_i} + \lambda \|\bfalpha_i\|_1
\vphantom{
\left.
-
\sum_{j=1}^N P(j=\mathcal{M}_i|\{\bfv_i\}) \times \mathcal{D}_{\bfv_i,\bfv_j}
\right]
}
\right .
\\
&
+\gamma
\left [
\sum_{j=1}^N P(j=\mathcal{H}_i|\{\bfv_i\})\times \mathcal{D}_{\bfv_i,\bfv_j}
\right .
\\
&
\left.
\left.
-
\sum_{j=1}^N P(j=\mathcal{M}_i|\{\bfv_i\}) \times \mathcal{D}_{\bfv_i,\bfv_j}
\right]
\right \},
\end{aligned}
\end{equation}
where the $\|\bfalpha_i\|_1$ is the $l_1$ norm based sparse term on $\bfalpha_i$,
and $\lambda$ and $\gamma$ are the trade-off weights, which are set by cross-validation.

\subsection{Optimization}

We adopt the EM framework to optimize the objective function
in (\ref{equ:objective}) in an iterative algorithm.
The algorithm is composed of two iterative steps: the E-step and the M-step.

\subsubsection{E-step}

In the E-step, we compute the probabilities of $P(j=\mathcal{H}_i|\{\bfv_i\})$ and $P(j=\mathcal{M}_i|\{\bfv_i\})$
based on $\{\bfv_i\}$ estimated in the previous iteration, as in (\ref{equ:probality}).

\subsubsection{M-Step}

In the M-Step, we try to optimize (\ref{equ:objective}) by fixing the probabilities.
By denoting
$P_{ij}=P(j=\mathcal{H}_i|\{\bfv_i\}) - P(j=\mathcal{M}_i|\{\bfv_i\})$,
the objective function in (\ref{equ:objective}) is reduced to
\begin{equation}
\label{equ:objective1}
\begin{aligned}
\underset{\{\bfalpha_i,\bfv_i\}}{min}
&\sum_{i=1}^N
\left(
\mathcal{R}_{\bfv_i,\bfalpha_i} + \lambda \|\bfalpha_i\|_1
+
\gamma \sum_{j=1}^N P_{ij} \mathcal{D}_{\bfv_i,\bfv_j}
\right).
\end{aligned}
\end{equation}
We optimize the representation parameters $\bfalpha_i$ and $\bfv_i$ for each set one by one.
To optimize $\bfalpha_i$ and $\bfv_i$, we fix all the remaining
representation parameters $\bfalpha_j$ and $\bfv_j(j \neq i)$.
Using the property $\mathcal{D}_{\bfv_i,\bfv_i}=0$ and $\mathcal{D}_{\bfv_i,\bfv_j}=\mathcal{D}_{\bfv_j,\bfv_i}$,
we rewrite the optimization of
objective of (\ref{equ:objective1}) with respect to only $\bfalpha_i$ and $\bfv_i$ as follows:
\begin{equation}
\label{equ:objective2}
\underset{\bfalpha_i,\bfv_i}{min}~
\left \{
\begin{aligned}
&H(\bfalpha_i,\bfv_i)\\
&=
\mathcal{R}_{\bfv_i,\bfalpha_i} + \lambda \|\bfalpha_i\|_1\\
&~~~~~~~+
\gamma
\left (
\sum_{j:j\neq i} P_{ij} \mathcal{D}_{\bfv_i,\bfv_j}
+
\sum_{j:j\neq i} P_{ji} \mathcal{D}_{\bfv_j,\bfv_i}
\right )
\\
&=
\mathcal{R}_{\bfv_i,\bfalpha_i} + \lambda \|\bfalpha_i\|_1+
\gamma
\sum_{j:j\neq i} (P_{ij}+P_{ji}) \mathcal{D}_{\bfv_i,\bfv_j}
\\
&=f(\bfalpha_i,\bfv_i) + g(\bfalpha_i)
\end{aligned}
\right \}
\end{equation}
The objective function $H(\bfalpha_i,\bfv_i)$ in (\ref{equ:objective2})
is a composite model consisting
of a smooth function
$f(\bfalpha_i,\bfv_i) = \mathcal{R}_{\bfv_i,\bfalpha_i} +
\gamma
\sum_{j:j\neq i} (P_{ij}+P_{ji}) \mathcal{D}_{\bfv_i,\bfv_j}$ and a non-smooth function
$g(\alpha_i) = \lambda \|\bfalpha_i\|_1$.

To solve the optimization problem in (\ref{equ:objective2}),
we employ the accelerated proximal
gradient (APG) algorithm introduced in \cite{Beck2009}.
In the $t$-th iteration,
to obtain the new solution $\bfalpha_i^t$ and $\bfv_i^t$, we solve
the following proximal regularization problem based on the previous
the solution $\bfalpha_i^{t-1}$ and $\bfv_i^{t-1}$:

\begin{equation}
\label{equ:objective3}
\bfalpha^t_i,\bfv_i^t=
\arg\min_{\bfalpha_i,\bfv_i}
\left \{
\begin{aligned}
&Q_{L^t}(\bfalpha_i,\bfbeta_i^t,\bfv_i,\bfu_i^t)\\
&=
\frac{L^t}{2}
\left \|
\bfalpha_i -
\left ( \bfbeta_i^t
-
\frac{1}{L^t}
\nabla f_{\bfbeta_i^t}
\right )
\right \|^2_2\\
&+
\frac{L^t}{2}
\left \|
\bfv_i -
\left ( \bfu_i^t
-
\frac{1}{L^t}
\nabla f_{\bfu_i^t}
\right )
\right \|^2_2\\
&+
g(\bfalpha_i)
\end{aligned}
\right \},
\end{equation}
where
\begin{equation}
\label{equ:beta}
\begin{aligned}
&\bfbeta_i^{t}=\bfalpha_i^{t-1}+
\frac{\tau^{t-1}-1}{\tau^t}\left(
\bfalpha^{t-1}_i-\bfalpha^{t-2}_i
\right ),\\
&\bfu_i^{t}=\bfv_i^{t-1}+
\frac{\tau^{t-1}-1}{\tau^t}\left(
\bfv^{t-1}_i-\bfv^{t-2}_i
\right ),
\end{aligned}
\end{equation}
with
$\tau^t=
\frac{1+
\sqrt{1+4 {\tau^{t-1}}^2}
}{2}
$
being the initial approximation of the next solution for $\bfalpha^t_i$,
$L^t
=
\eta^{i^t}L^{t-1}$  being the step size related to the Lipschitz constant,
and $i^t$ being the smallest nonnegative integers  such that
$H(\bfalpha_i^{t-1},\bfv_i^{t-1})\leq Q_{L^t}(\bfalpha_i^{t-1},\bfbeta_i^t,
\bfv_i^{t-1},\bfu_i^t)$,
and
\begin{equation}
\begin{aligned}
\nabla f_{\bfbeta_i^t}
=&
\left.
\frac{\partial f(\bfv_i,\bfalpha_i)}{\partial \bfalpha_i}
\right |_{\bfalpha_i=\bfbeta_i^t}
=-2 X_i^\top (\mu_i+U_iv_i - X_i \bfbeta_i^t),\\
\nabla f_{\bfu_i^t}
=&
\left.
\frac{\partial f(\bfv_i,\bfalpha_i)}{\partial \bfv_i}
\right |_{\bfv_i=\bfu_i^t}\\
=&
2 \left[1+\gamma \sum_{j:j\neq i} (P_{ij}+P_{ji})\right]
U_i^\top (U_i \bfu_i^t+\bfmu_i)\\
&-2U_i X_i \bfalpha_i - 2 \gamma \sum_{j:j\neq i} (P_{ij}+P_{ji}) (\bfmu_j-U_j \bfu_j^t).
\end{aligned}
\end{equation}
It could be proved that
when $g(\bfalpha_i)$ is given as $g(\bfalpha_i)=\lambda \|\bfalpha_i\|_1$,
the optimal solution of (\ref{equ:objective3}) can be obtained by

\begin{equation}
\label{equ:alpha}
\begin{aligned}
\bfalpha^t_i
&=
\tau_{
\frac{\lambda}{L^t}}
\left(
\bfbeta_i^t
-
\frac{1}{L^t}
\nabla f_{\bfbeta_i^t}
\right),\\
\bfv^t_i
&=
\bfu_i^t
-
\frac{1}{L^t}
\nabla f_{\bfu_i^t},
\end{aligned}
\end{equation}
where $\tau_{\varepsilon}(w)=
\left(
|w|-\varepsilon
\right)_+
sgn(w)
$
is the soft-thresholding
operators.
The final iterative representation algorithm for LaMa-IS method is summarized in
Algorithm
\ref{alg:LaMaIS-Rep}.

\begin{algorithm}[h!]
\caption{Large Margin Image Set Representation Algorithm (LaMaIS-Rep)}
\label{alg:LaMaIS-Rep}
\begin{algorithmic}
\STATE \textbf{Input}: Training Image Sets $\{(X_i,y_i)\}_{i=1}^N$;
\STATE \textbf{Input}: Maximum iteration number $T$;
\STATE Initialize the representation parameters  $\{(\bfalpha_i^0,\bfv_i^0)\}_{i=1}^N$,
$\{(\bfalpha_i^{-1},\bfv_i^{-1})\}_{i=1}^N$ and $\tau^0$;

\FOR{$t=1,\cdots,T$}

\FOR{$i=1,\cdots,N$}

\STATE Update the probabilities of $P(j=\mathcal{H}_i|\{\bfv_i^{t-1}\})$ and $P(j=\mathcal{M}_i|\{\bfv_i^{t-1}\})$ as in (\ref{equ:probality});

\STATE Update the approximated initial representation parameters $\bfbeta_i^{t}$ and $\bfu_i^{t}$ as in (\ref{equ:beta}).

\STATE Update the representation parameters $\bfalpha_i^{t}$ and $\bfv_i^{t}$ as in (\ref{equ:alpha}).

\ENDFOR
\ENDFOR

\STATE \textbf{Output}: Optimal solution $\{(\bfalpha_i^{T},\bfv_i^{T})\}_{i=1}^N$.

\end{algorithmic}
\end{algorithm}

\subsection{Classification}

Given a test image set $X_k$, we first
assume that it belongs to class $y\in\{1,\cdots,C\}$,
and then compute the distance to the sets of class $y$
as
\begin{equation}
\begin{aligned}
\mathcal{E}_y=
\underset{\bfalpha_k,\bfv_k}{\min}
&
\left \{
\mathcal{R}_{\bfalpha_k,\bfv_k}
+
\lambda \|\bfalpha_k\|_1
\vphantom{
\left .
\sum_{j=1}^N
P(i=\mathcal{M}_k^{y}|\{\bfv_i\},\bfv_k)\times \mathcal{D}_{\bfv_k,\bfv_i}~
\right ]
}
\right.\\
&+
\gamma
\left [
\sum_{j=1}^N
P(i=\mathcal{H}_k|\{\bfv_i\},\bfv_k,y_k=y)\times \mathcal{D}_{\bfv_k,\bfv_i}
\right .\\
&
-
\left .
\left .
\sum_{j=1}^N
P(i=\mathcal{M}_k|\{\bfv_i\},\bfv_k,y_k=y)\times \mathcal{D}_{\bfv_k,\bfv_i}~
\right ]
\right\},
\end{aligned}
\end{equation}
where
$P(i=\mathcal{H}_k|\{\bfv_i\},\bfv_k,y_k=y)$
is the probability of $X_i$ being the nearest hit of $X_i$
conditional on $y_k=y$.
This problem can also
be solved by an EM algorithm with the APG optimization.
We then assign
a label $y_k$ to $X_k$ as follows:
\begin{equation}
\begin{aligned}
y_k=
\underset{y\in \{1,\cdots,C\}}{argmin}~ \mathcal{E}_y
\end{aligned}
\end{equation}

\subsection{Time Complexity}

Given the iteration number $T$,
in the off-line procedure of our method,
the
presentation parameters are updated one by one for $T$ times.
Since there are $N$ training image sets, the training time complexity is $O(N\times T)$.
In the on-line classification procedure,
the test image set is compared to each class, and in each comparison,
the representation parameter of the test image set is updated for $T$ times,
and the representation parameters of the training image set are fixed.
Since the number of classes is $C$, the time complexity for the on-line classification is
$O(C\times T)$.
In contrast, the SANP algorithm does not have a training procedure.
The test image set is compared to all the training image sets.
In each comparison,  the parameter of both
the test image and training image set   are updated, thus the time complexity
is $O(2\times N\times T)$.
Usually $C\ll N$, thus the time complexity is reduced significantly compared to SANP.

\section{Experiments}
\label{sec:experiment}

We evaluated the proposed method on two comprehensive face image set classification tasks where the goal is to conduct video-based face recognition.

\subsection{Dataset and Setup}

To evaluate our method, we used two real-world face video sequence datasets:

\begin{itemize}
\item \textbf{Conference Face Set}:
We collected a human face video sequence dataset from an
international academic conference of 5 days.
The videos of 32 different conference participants' faces are captured during the oral session, poster session,
reception banquet and some other casual scenes with an handheld camera.
The face images extracted from the frames of this dataset cover large variations in illumination, head pose and
facial expression, making the face recognition task from these videos quite challenging.
For example, the oral presenters' faces were captured during both oral presentation procedure when light was off and
Q\&A procedure when light was on.
Moreover, the participants' expressions were quite serious when discussing academic problems, while
much more relaxed during the reception banquet.
All the images are taken from the same coordinates, but not always of the same sample size.
The videos of 32 participants were first captured and then split into totally 507 continuous video sequences to construct
the dataset.
The number of video sequences for each participant varies from 12 to 19.
The entire dataset was further divided into a training set (328 sequences) and an independent test set (179 sequences).
The statistic information of the dataset is summarized in Table \ref{tab:ConfData}.
The frames of each sequence are considered as an image set, and
the face region is firstly extracted, then
the local binary pattern
(LBP) features are extracted from the face region, and finally the LBP feature vectors
are combined to construct the
data matrix of an image set.

\begin{table*}[htb!]
\centering
\caption{Statistics of the conference face image set database.}
\label{tab:ConfData}
\begin{tabular}{|l|c|c|c|}
\hline
Sets & Sample Number &  Participant Number &Samples for Each Participant\\
\hline \hline
Training Set & 328 & 32& 8  $\sim$ 13\\
Test Set & 179 &32& 3  $\sim$ 7\\
\hline
Entire Set & 507 &32& 12  $\sim$ 19\\
\hline
\end{tabular}
\end{table*}

\item \textbf{YouTube Face Set}:
We also used a large-scale face image set database ---
YouTube Celebrities database \cite{Kim2008}.
It is the  largest video
database proposed for video face tracking and recognition.
In this database,
1910 video sequences
of 47 celebrities
(actors, actresses and politicians) were collected
from YouTube.
The number of sequences for each celebrity varies from 17 to 108.
The number of frames of each sequence varies from 8 to 400,
and most frames are of low
resolution and highly compressed.
This database is more challenging than the other
image set databases, due to the large
variations
in poses, illuminations and expressions.
We also randomly split the database into a training set (1275 sequences) and
an independent test set (635 sequences).
The statistics of the training set and test set is given in
Table \ref{tab:YoutubeData}.
The face area is cropped from each frame and scaled to size of $20\times 20$,
and the pixel values are used as the visual features after histogram equalization.

\begin{table*}[htb!]
\centering
\caption{Statistics of the Youtube face image set database.}
\label{tab:YoutubeData}
\begin{tabular}{|l|c|c|c|}
\hline
Sets & Sample Number &  Person Number &Samples for Each Person\\
\hline \hline
Training Set & 1910& 47 & 11 $\sim$ 72\\
Test Set & 635 & 47 & 6 $\sim$26\\
\hline
Entire Set & 1910& 47 & 17 $\sim$ 108\\
\hline
\end{tabular}
\end{table*}
\end{itemize}

To conduct the experiments, we first performed the cross validation to the training sets
to select the optimal parameters.
The 8-fold cross validation and 10-fold cross validation were performed to
the conference face database and the
YouTube face database, respectively.
Using the parameters learned by the training set,
we classified the image sets in the independent test sets.
The classification accuracies are reported as the
performance measure.

\subsection{Results}

\begin{figure*}[!htb]
\centering
\subfigure[8-fold cross validation on the training set]{
\includegraphics[width=0.45\textwidth]{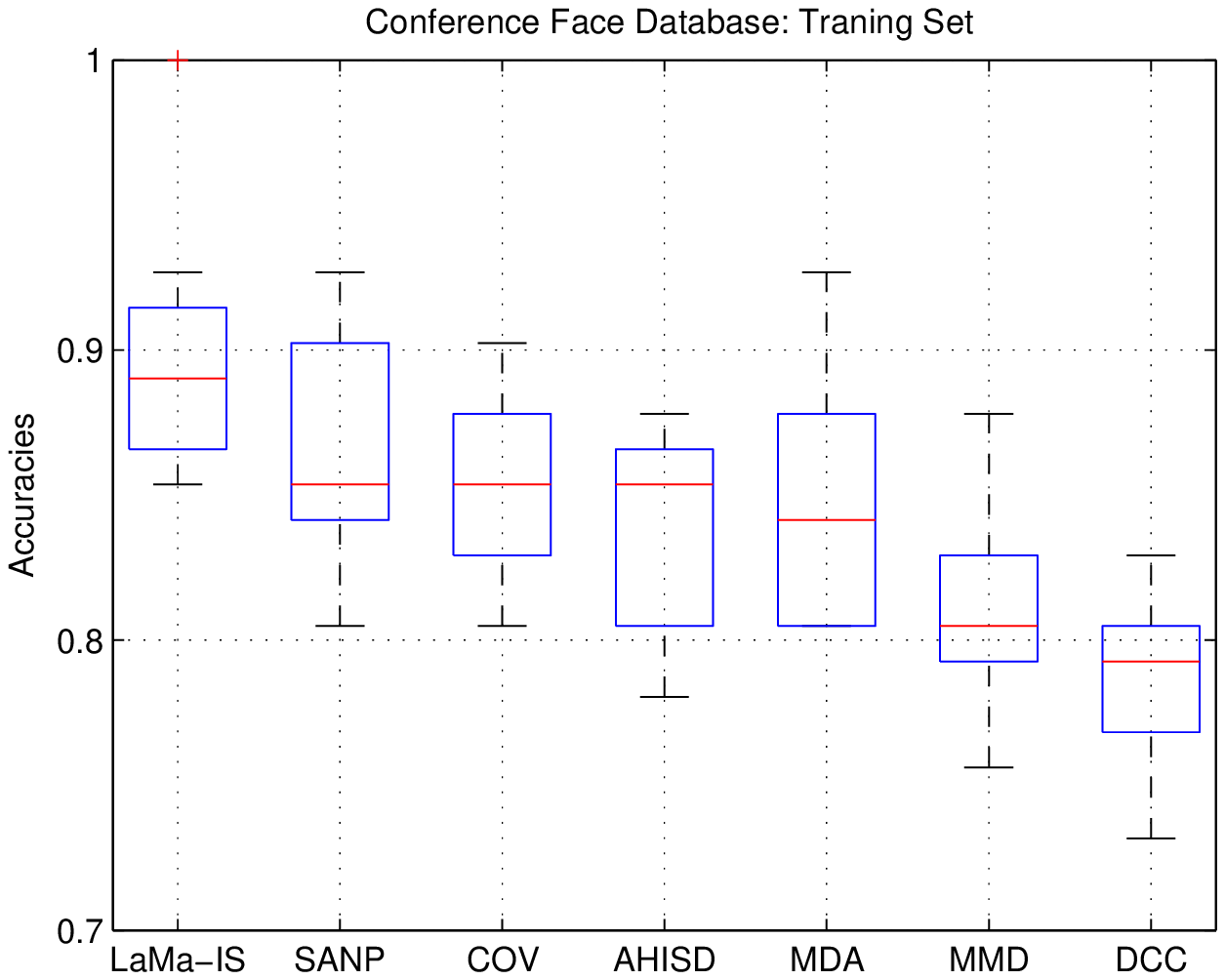}}
\subfigure[Evaluation on the independent test set]{
\includegraphics[width=0.45\textwidth]{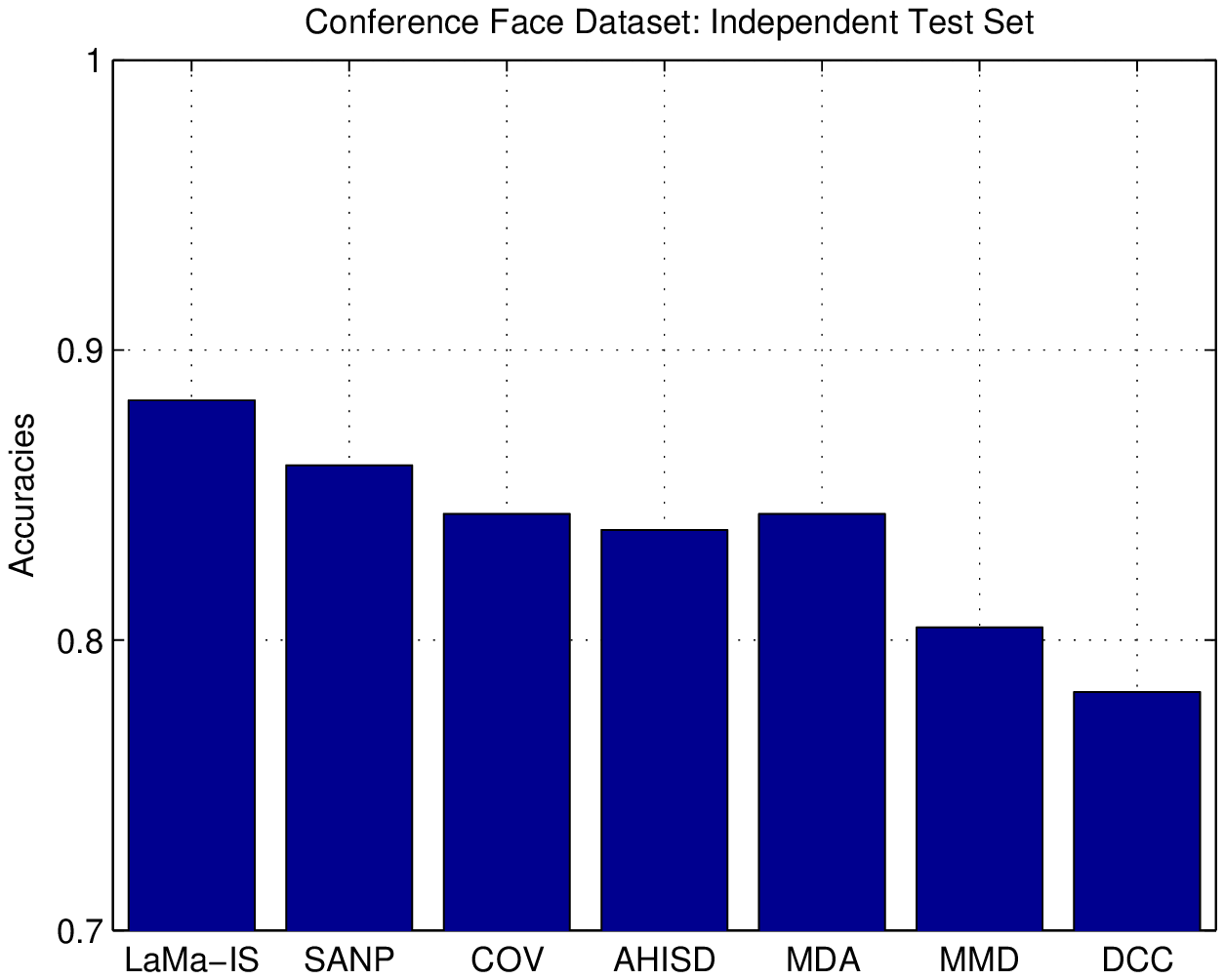}}\\
\caption{Performance of different methods on the conference face image set database.}
\label{fig:FigConf}
\end{figure*}

We compared our method against the state-of-the-art image set classification methods, including
DCC \cite{Kim2008}, MMD \cite{MMD2008}, MDA \cite{MDA2009}, AHISD \cite{AHISD2010},
COV \cite{Wang2012}, and
SANP \cite{Hu2011,Hu2012}.
The performance of different methods on the conference face database is shown in Figure \ref{fig:FigConf}.
In Figure \ref{fig:FigConf} (a), the
boxplots of the accuracies of  the 8-fold cross validation is shown,
while in  Figure \ref{fig:FigConf} (b),
the accuracies of evaluation on  the independent test set is given.
It is obvious that LaMa-IS outperforms all other methods
on both the training and test sets.
The better performance of LaMa-IS is
mainly due to the
usage of both the image set class labels
and the exploration of the  global structure of the
image set database.
MDA also matches the image samples against its neighbors of the same set and different sets, which
is another strategy of the large margin framework.
However, we defined the margin at the image set level, instead of the individual image sample level.
Thus although MDA also archives good classification accuracy,
LaMa-IS performs significantly better than MDA.
SANP and LaMa-IS both use the same two models
to represent the image sets, but SANP only focuses on the comparison of a pair of image sets,
whereas LaMa-IS learns the representations for all the training
image sets in a discriminant manner.
It turns out that tuning the representation parameters coherently for all the image sets
by using the class labels is not a trivial task, which can significantly improve the performance.
This can be verified by the outperformance of LaMa-IS over SANP.
COV represents each image set as a covariance matrix and also utilizes
the class labels to learn the representation parameters.
However, its performance is inferior to both LaMa-IS and SANP,
which means that the representation model of using both image samples
and affine hull model is more effective than the covariance matrix,
especially when the training sample number is not large.

\begin{figure*}[!htb]
\centering
\subfigure[10-fold cross validation on the training set]{
\includegraphics[width=0.45\textwidth]{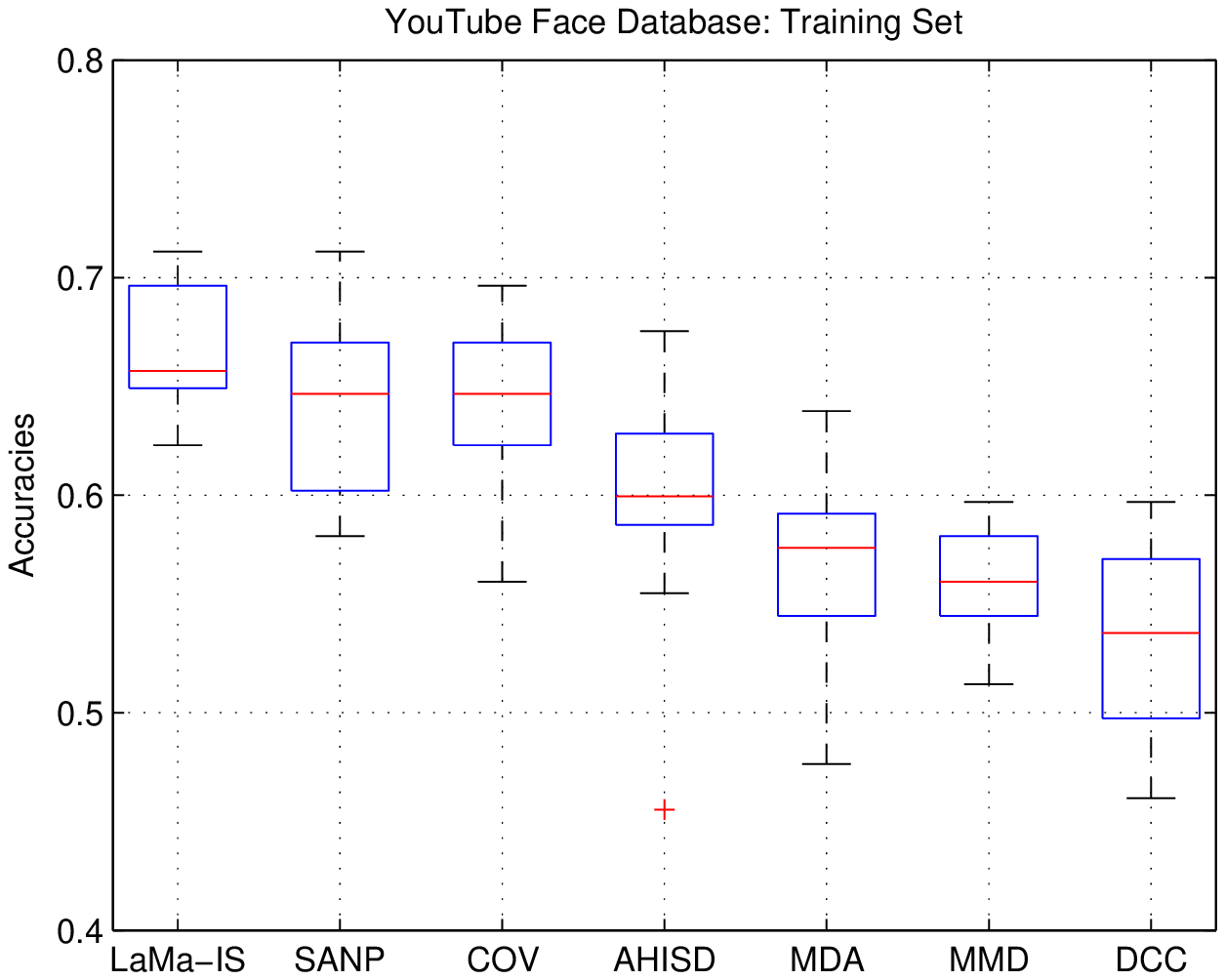}}
\subfigure[Evaluation on the independent test set]{
\includegraphics[width=0.45\textwidth]{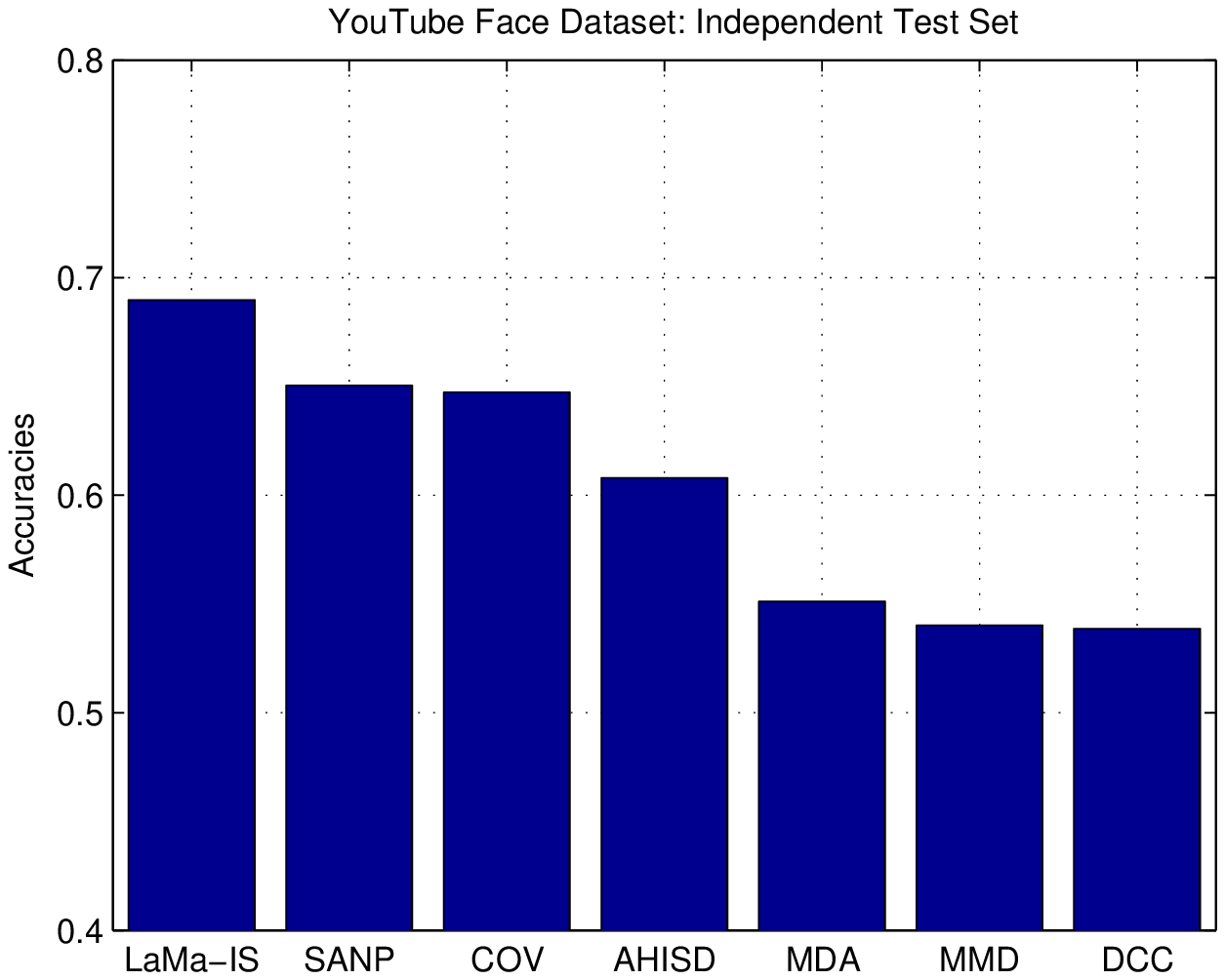}}\\
\subfigure[Computation time]{
\includegraphics[width=0.45\textwidth]{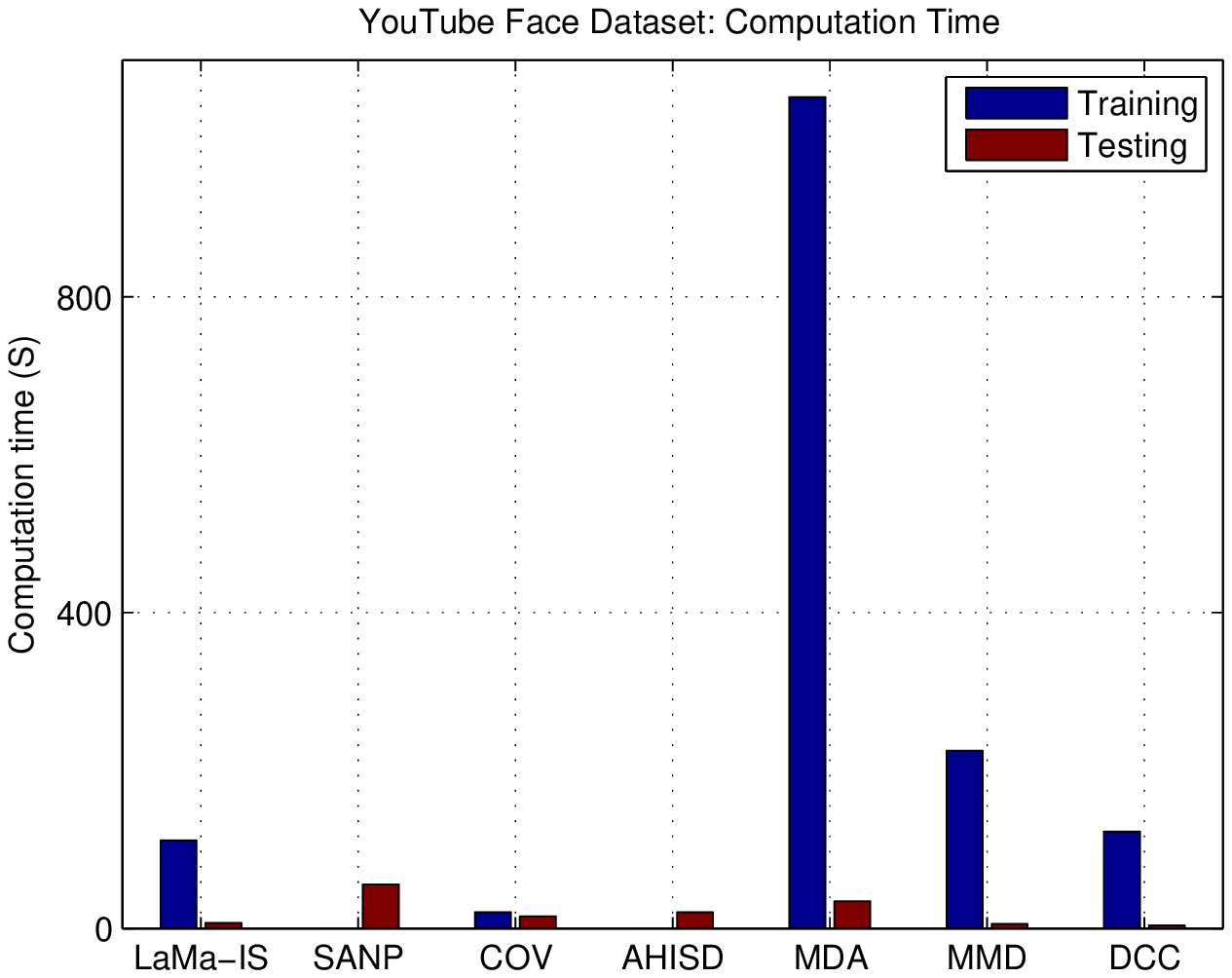}}\\
\caption{Runtime of different methods on the YouTube face image set database.}
\label{fig:FigYoutube}
\end{figure*}

Figure \ref{fig:FigYoutube} shows the performance of different methods on the YouTube face set.
It can be seen that LaMa-IS also outperforms all other methods, on both the training set and test set.
Different from the results in Figure \ref{fig:FigConf},
COV outperforms SANP slightly.
The possible reason is that the YouTube database provides more
image sets and thus provides sufficient discriminant information to be utilized by COV.

We further compared the computation time
of different methods on the YouTube database for training and testing (classification of one image set),
and reported the results in Figure \ref{fig:FigYoutube} (c).
It could be seen that the training procedure for MDA, MMD and DCC is quite time consuming,
and at the same time, the accuracy of these three methods is not satisfactory.
SANP, on the other hand, has quite high time cost on the on-line testing process, but requires no time on training as it does not require a training step. Its performance is also much better than MDA, MMD and DCC.
Comparing with SANP, the proposed LaMa-IS method requires an off-line training procedure, which does not consume much time,
but can boost the classification performance significantly on this large scale database.
Additionally, its on-line classification procedure is also computationally efficient.

\section{Conclusion}
\label{sec:conclusion}

In this paper, we have proposed a novel large-margin-based image set representation and classification method, LaMa-IS.
To represent an image set, LaMa-IS encodes information from both the image samples in the set and their affine hull models. We defined the margin of an image set as
the difference of the distance to its nearest
image set from different classes
and the distance to its nearest image set of the same class.
The maximum margin is optimized by an expectation---maximization (EM) strategy
with accelerated proximal
gradient (APG) optimization
in an iterative algorithm.
In the classification procedure, LaMa-IS compares a test image set to every class, instead of every image set,
making it a computationally efficient algorithm for large-scale applications.
Experimental results on two comprehensive face image set classification tasks demonstrate that the proposed method significantly outperforms the state-of-the-art methods in terms of both effectiveness and efficiency.


\end{document}